\documentclass[sigconf]{acmart}

\AtBeginDocument{%
    \providecommand\BibTeX{{%
        \normalfont B\kern-0.5em{\scshape i\kern-0.25em b}\kern-0.8em\TeX}
    }
}

\setcopyright{acmcopyright}
\copyrightyear{2023}
\acmYear{2023}
\acmDOI{XXXXXXX.XXXXXXX}

\acmConference[Conference acronym 'XX]{Make sure to enter the correct
  conference title from your rights confirmation emai}{June 03--05,
  2018}{Woodstock, NY}

\acmPrice{15.00}
\acmISBN{978-1-4503-XXXX-X/18/06}

\usepackage{enumitem}
\usepackage{framed} 

\usepackage{url}
\usepackage{hyperref}
\usepackage[hyphenbreaks]{breakurl}

\begin{document}

\title{Computational analyses of linguistic features with schizophrenic and autistic traits along with formal thought disorders}
\renewcommand{\shorttitle}{Computational analyses of linguistic features with schizophrenic and autistic traits along with FTDs}

\author{Takeshi Saga}
\affiliation{%
  \institution{NAIST}
  \streetaddress{8916-5 Takayama-cho}
  \city{Ikoma}
  \state{Nara}
  \country{Japan}
  \postcode{630-0192}
}
\email{saga.takeshi.sn0@is.naist.jp}

\author{Hiroki Tanaka}
\affiliation{%
  \institution{NAIST}
  \streetaddress{8916-5 Takayama-cho}
  \city{Ikoma}
  \state{Nara}
  \country{Japan}
  \postcode{630-0192}
}
\email{hiroki-tan@is.naist.jp}

\author{Satoshi Nakamura}
\affiliation{%
  \institution{NAIST}
  \streetaddress{8916-5 Takayama-cho}
  \city{Ikoma}
  \state{Nara}
  \country{Japan}
  \postcode{630-0192}
}
\email{s-nakamura@is.naist.jp}

\renewcommand{\shortauthors}{Saga, et al.}

\begin{abstract}

    Formal Thought Disorder (FTD), which is a group of symptoms in cognition that affects language and thought, can be observed through language. 
    FTD is seen across such developmental or psychiatric disorders as Autism Spectrum Disorder (ASD) or Schizophrenia, and its related Schizotypal Personality Disorder (SPD). 
    Researchers have worked on computational analyses for the early detection of such symptoms and to develop better treatments for more than 40 years. 
    This paper collected a Japanese audio-report dataset with score labels related to ASD and SPD through a crowd-sourcing service from the general population. 
    We measured language characteristics with the 2nd edition of the Social Responsiveness Scale (SRS2) and the Schizotypal Personality Questionnaire (SPQ), including an \textit{odd speech} subscale from SPQ to quantize the FTD symptoms. 
    We investigated the following four research questions through machine-learning-based score predictions: 
    (RQ1) How are schizotypal and autistic measures correlated?
    (RQ2) What is the most suitable task to elicit FTD symptoms?
    (RQ3) Does the length of speech affect the elicitation of FTD symptoms?
    (RQ4) Which features are critical for capturing FTD symptoms?
    We confirmed that an FTD-related subscale, \textit{odd speech}, was significantly correlated with both the total SPQ and SRS scores, although they themselves were not correlated significantly. 
    In terms of the tasks, our result identified the effectiveness of FTD elicitation by the most negative memory. 
    Furthermore, we confirmed that longer speech elicited more FTD symptoms as the increased score prediction performance of an FTD-related subscale \textit{odd speech} from SPQ. 
    Our ablation study confirmed the importance of function words and both the abstract and temporal features for FTD-related \textit{odd speech} estimation. 
    In contrast, content words were effective only in the SRS predictions, and content words were effective only in the SPQ predictions, a result that implies the differences between SPD-like and ASD-like symptoms. 
    Data and programs used in this paper can be found here: \url{https://sites.google.com/view/sagatake/resource}.

\end{abstract}




\maketitle

\section{Introduction}

    Formal Thought Disorder (FTD) is a group of symptoms seen in most psychiatric disorders and people with clinical high risks \cite{TALD}. 
    FTD is a disorder of the form of thought (e.g., pressured speech) rather than its content (e.g., delusions). 
    %
    %
    Since the '80s, researchers have conducted FTD research on schizophrenia populations since FTD is their primary symptom \cite{Manschreck1981,Kircher2018}. 
    However, subsequent research concluded that it is also seen in other neuropsychiatric or developmental disorders, including mania or autism spectrum disorder (ASD) \cite{Yalicetin2017,Solomon2008}. 


    Toward automatic diagnosis and detailed understanding of disorders, automatic computational analyses are increasingly being utilized to investigate psychiatric disorders based on the evolution of natural language processing (NLP) techniques. 
    For example, Linguistic Inquiry and Word Count (LIWC) is a common computational software that can automatically calculate such linguistic statistics as part-of-speech and total word count \cite{LIWC2022}. 
    Silva et al. compared a value called \textit{Analytic Thinking} (originally \textit{Categorical-Dynamic Index (CDI)}) \cite{Pennebaker2015} from LIWC with clinical FTD measures: Positive and Negative Syndrome Scale-8 Item (PANSS-8) and Thought Language Index (TLI) \cite{Silva2021}. 
    They showed the usefulness of analytic thinking scores for predicting FTD scores, where no single word type drove the FTD scores well. 
    Similar to methods based on descriptive statistics (e.g., LIWC), embedding-based NLP methods are promising options for computational analyses. 
    In the context of schizophrenia analysis, Tang et al. showed the characteristic difference of embedding features derived from bidirectional encoder representation from transformers (BERT) between schizophrenia and control groups \cite{Tang2021}. 
    By comparing the values of \textit{the assessment of Thought, Language, and Communication (TLC)} and the BERT-derived features between the schizophrenia and control groups, they found that only the former showed an increase of an absolute difference of embedding vectors in the sentence level along with an increase of the number of response sentences for interviewer questions. 
    In addition to these two approaches, other research uses such variations of NLP techniques as graph-based approaches and approaches that use referential nouns \cite{Mota2012,Cokal2022}.

    Although both of these papers showed the effectiveness of features for FTD analyses, comparing these methods is difficult since their settings are different in the following ways. 
    One is task setting. 
    Since FTD symptoms can be seen in a variety of stimulation, including image descriptions (e.g. thematic apperception test), clinical interviews, and memory reports \cite{Silva2021,Tang2021,Mota2012}. 
    Another is speaking duration. 
    Some research used only 30 seconds of audio; others used 12 minutes. 
    Since machine-learning approaches are sensitive to the data amount, a valid comparison requires duration alignment. 
    In addition to these setting alignment problems, differences in linguistic features between schizophrenic and other symptomatic characteristics remain under-investigated areas. 
    To better understand FTD symptoms across different psychiatric aspects, quantitative analysis is required. 
    The symptomatic similarities between schizophrenia and ASD are exciting research targets.
    Since recent papers showed symptomatic overlaps and differences between Schizophrenia and ASD with various clinical assessments, we set these symptoms as targets in this study \cite{Konstantareas2001,Foss-Feig2016,SCI-PANSS}. 
    As a feasibility study, we collected and analyzed participants from general populations by psychiatric personality scores. 
    In this paper, we collected various conditioned audio-report datasets with schizotypal/autistic labels and analyzed their characteristics with computational approaches to answer the following four research questions: 
    \begin{enumerate}
        \item[(RQ1)] How are schizotypal and autistic measures correlated?
        \item[(RQ2)] What is the most suitable task for capturing FTD symptoms?
        \item[(RQ3)] Does speech length affect the elicitation of FTD symptoms?
        \item[(RQ4)] Which features are important for capturing FTD symptoms?
    \end{enumerate}

    Our paper has the following scientific novelties: (a) We showed the effectiveness of negatively perceived elicitation by comparing the negative and the mistake memory-reminding tasks. 
    (b) We investigated speech duration effects on FTDs, which revealed that longer speech is superior in eliciting FTD symptoms. 
    (c) We compared several NLP methods’ characteristics with the same dataset.


\section{Dataset}

    \subsection{Data collection}
    
        We collected an audio-report dataset with schizotypal and autistic personality scores. 
        Inspired by Mota's work \cite{Mota2017}, we compared task differences using reports about a recent dream or a favorite thing, a negative memory, or one's biggest mistake.
        For reliable machine-learning results, we used a crowd-sourcing approach to collect the largest possible dataset. 
        Although previous work used affective image elicitation, using such images online was prohibited by the provider \cite{Mota2017,IAPS}. 
        Therefore, we replaced them with topic sentences. 
        To ensure that the negative affect elicited FTD symptoms, we added one's biggest mistake as a task because it resembles the negative memory task, although it is a more fact-driven topic. 
    
        First, we got informed consent from the participants concerning their data usage and then conducted psychiatric questionnaires. 
        We used Japanese versions of the following questionnaires: the Schizotypal Personality Questionnaire (SPQ) for schizotypal symptoms \cite{SPQ}; the 2nd edition of the Social Responsiveness Scale (SRS2) for autistic symptoms \cite{SRS_book}; Kikuchi's Social Skills Scale of 18 items (KiSS18) for social skills \cite{KiSS-18}; the Ten Item Personality Inventory (TIPI) for general personalities \cite{TIPI}. 
        We measured the symptomatic characteristics of the participants with SPQ and SRS2. 
        SPQ is a questionnaire on schizotypal personalities with 74 yes/no questions; each scored answer as 1 for “yes” and 0 for “no.” 
        It also provides subscales, including “odd speech,” which is almost identical to formal thought disorders (FTDs), constructed with nine questions. 
        Therefore, the total score range and the odd speech range are 0-74 and 0-9, respectively. 
        Unlike such face-to-face clinical assessments as TLI \cite{TLI} or TALD \cite{TALD}, we utilized questionnaire-style measures to quantify their characteristics through a crowd-sourcing service. 
        SPQ is especially suitable for this research since it can calculate an FTD-related subscale named \textit{odd speech}. 
        Similarly, SRS2 is a questionnaire on autistic traits with 65 items in 4 point Likert scale. 
        We measured both of SPQ and SRS2 to compare schizophrenic and autistic FTD-like symptoms quantitatively. 
        %
        Second, we asked them to talk and record their voices on the following topics: a recent dream, a negative memory, and their biggest mistake. 
        If they couldn't remember a recent dream, we asked them to talk about their favorite things corresponding to the positive images used in Mota's work. 
        We limited the duration of each recording to 30, 60, or 120 seconds.
        The recording started with a recent dream or their favorite thing with a 30-second limit. 
        Then the recording scenario addressed their negative memory for 30 seconds and another 30 seconds for their biggest mistake. 
        After they finished their recordings for every task with a 30-second limit, they recorded identical tasks but with two different time limits: 60 and 120 seconds. 
        In total, each participant recorded 9 sessions. 
        All the recordings were done through web-based software ({\url{https://online-voice-recorder.com/ja/}) to attenuate any software differences. 
        Since the software saves the audio onto their local hard drives, they upload their files to our cloud storage. 
        We gave these instructions using Google Forms. 
        %
        %
        We transcribed all the audio with a pretrained automatic speech recognition (ASR): Whisper-large-v2 \cite{Whisper}. 
        Since Whisper recognized and punctuated their transcripts, we can assess pseudo pauses using only text transcripts without complex audio processing. 
        Although our web-based procedure contained a potential limitation of uncontrollable environmental settings (e.g., microphone gain, background noises, recording rooms), we standardized this situation using state-of-the-art ASR-powered audio transcription, machine-learning models, and a large amount of data. 
        We collected audio files from 54 participants from 446 sessions: 58 on dream reports, 91 on favorite things, 149 on negative memories, and 148 on biggest mistakes.

        Table~\ref{tab:observed_stat} shows the observed distribution of the score labels and demographic information. 
        Since we recruited participants through an anonymized crowd-sourcing service, we only collected limited demographic information to calculate the SRS cutoff (age and gender). 
        Although it is better to recruit actual patients, we started from the general population using a crowd-sourcing service as exploratory research to validate methodological appropriateness. 
        We chose 81 and 41 as the cutoff values for SRS and SPQ, and 28 participants (36.84$\%$ of all) and 15 participants (17.56$\%$) were potential candidates for ASD and SPD, respectively \cite{SPQ,SRS_distribution}. 
        Therefore, our dataset suits our initial investigation toward future experiments with diagnosed patients.

        \begin{table}[t]
            \centering
            \caption{\textbf{Observed distribution of the score labels and demographics}. Obs. range = observed range.}
            \label{tab:observed_stat}
            \begin{tabular}{||l||c|c||c|l||}
                \toprule
                            & Mean   & SD    & Misc.\\
                \midrule
                \midrule
                Age         & 43.22     & 11.36 & Obs. range: 19-74\\
                Gender      & -         & -     & Male: 42, Female: 50 \\
                \midrule
                SRS         & 73.27     & 28.90 & Obs. range: 2-152\\
                SPQ         & 28.41     & 16.06 & Obs. range: 0-74\\
                Odd Speech  & 3.88      & 2.86  & Obs. range: 0-9\\
                \bottomrule
            \end{tabular}
        \end{table}
        

    \subsection{Between-metric comparative analysis}

        We calculated Spearman's correlation coefficient to understand the similarities and differences in the characteristics of schizotypal and autistic measures as answers for RQ1. 
        Since SPQ includes general symptomatic characteristics of schizotypal personality disorders, we analyzed its total scores and subscales. 
        We set the following analysis axes: total scores and \textit{odd speech} from SPQ and total scores and \textit{awareness, cognition communication} from SRS2. 
        We additionally collected KiSS18 and TIPI-J for a general analysis of social skills and personalities for future research.
        
        \begin{table}[t]
            \centering
            \caption{\textbf{Spearman's correlation between schizotypal and autistic measures:} Bold indicates significant correlations in no-correlation test (p < 0.05) and Odd., Awr., Cog., Com. indicate \textit{odd speech} from SPQ, awareness, cognition, communication from SRS respectively.}
            \label{tab:metric_correlation}
            \begin{tabular}{||l||c|c||c|c|c|l||}
                \toprule
                            & SPQ       & Odd. & SRS & Awr. & Cog.  & Com. \\
                \midrule
                \midrule
                 SPQ        & 1.00      & \textbf{0.28}  & 0.01  & \textbf{-0.22} & \textbf{0.44}  & \textbf{0.28} \\
                 Odd &           & 1.00  & \textbf{0.32}  & \textbf{-0.29} & 0.10  & \textbf{0.11} \\
                \bottomrule
            \end{tabular}
        \end{table}

        Table~\ref{tab:metric_correlation} shows the result correlation between the metrics. 
        In terms of total scores, SPQ didn't correlate significantly with SRS.
        We confirmed that SPQ measures different clinical aspects than SRS at the abstract total score level.
        However, interestingly, SPQ showed significant correlations with the SRS subscales: awareness, cognition, and communication. 
       This result suggests that SPD and ASD symptoms are correlated in these aspects. 
        %
        
        The \textit{odd speech} was significantly correlated with both the total scores of SRS and SPQ. 
        This supports a previous finding in the comparative analysis of clinical measures between schizophrenia and ASD, suggesting symptomatic overlaps in negative symptoms (e.g., a flat affect and impoverished speech) \cite{Trevisan2020}.
        Interestingly, we captured the overlaps even though we used simpler questionnaire measures of SPD and ASD, which suggested the satisfying reliability of the questionnaires as the ground-truth labels for our analysis. 
        
\section{Experiment}

    \subsection{Settings}
    \label{sec:ML_setting}
    
        In this section, we investigated each effect by comparing the score estimation performances. 
        For the target scores, we focused on the total scores of SPQ and SRS and the \textit{odd speech} from SPQ as an FTD-specific measure.
        We used partial least square (PLS) regression as the machine-learning model, which eliminates inter-feature multicollinearities. 
        In addition to multicollinearity elimination, we removed the individual characteristic effects by leave-one-participant-out cross-validation. 
        Within each cross-validation loop, we used another five-fold cross-validation to optimize the number of principal components. 
        Although we assume that such other machine-learning approaches as lightGBM or XGBoost might improve score prediction performance, we chose this model because it provides stable performance with minimal hyperparameter tuning efforts to focus on our investigations of FTD-related phenomena. 
        
        We used the following five feature groups as input: embedding features, content word features, function word features, abstract features, and temporal features.
        %
        %
        The embedding features are composed of similarity scores between consecutive sentences or words based on BERT embeddings. 
        We used the approaches of Tang and Saga, which have confirmed their effectiveness in the schizophrenia and ASD analyses \cite{Tang2021,Saga2022_IEICE}. 
        Both approaches use a very similar algorithm. 
        First, they calculated the word or sentence embeddings with BERT. 
        Second, they calculated the similarity scores for each consecutive embedding.
        Third, they averaged and used the similarity scores as a feature value. 
        Tang's work used the L1-norm as a similarity measure at the sentence level. 
        In contrast, Saga used cosine distance at the sentence level or the content word level for their calculation. 
        We calculated the percentages of each part-of-speech (POS) tag divided into content or function word groups. 
        We used Sudachi for POS tagging \cite{Sudachi}.
        We selected nouns, verbs, adjectives, adjective verbs, and adverbs as content words and the rest as function words. 
        The abstract group includes a percentage of content words, the frequency of negation phrases, and a categorical-dynamic index (CDI, also known as analytic thinking in LIWC) \cite{Pennebaker2015,LIWC2022}. 
        Pennebaker originally proposed CDI for student essay analysis, and Silva et al. showed its effectiveness in schizophrenia classification \cite{Silva2021}. 
        Since CDI was originally developed for an English POS system, we adopted the procedure to the Japanese language and developed the Japanese version of CDI (CDIJ).
        First, we applied principal component analysis (PCA) to each POS feature. 
        Second, we picked plus/minus signs for each POS feature loading along with the 1st PCA component. 
        Third, we used the signs to calculate the CDIJ score as follows: 
        \begin{math}
            CDIJ=+[prefix]+[suffix]+[interjection]+[aux. verb]-[pronoun]-[adnominal]-[conjunction]-[particle]-[negation phrase]
        \end{math}.
        Unlike English, Japanese doesn't have specific negation words or phrases; instead, we used negation phrases detected with the KNP library \cite{KNP}.
        For temporal features, we included word per minute (WPM) and the punctuation percentage. 
        Since Whisper also estimates punctuation symbols based on vocal waveforms or pauses, we assume a pseudo-pause frequency. 
        Although researchers showed the usefulness of LIWC in psychiatric analyses, its Japanese version had just been published in 2022 \cite{J-LIWC}. 
        Therefore, we did not use it in this paper.

    \subsection{Task comparison}
    
        To identify suitable tasks regarding RQ2, we compared the performances among the four tasks to find suitable tasks to elicit FTD symptoms. 
        \begin{table}[t]
            \centering
            \caption{
                \textbf{Task comparison: Spearman's correlation between ground-truth and predictions:} Bold indicates significant correlations in the no-correlation test (p < 0.05).
            }
            \label{tab:task_comparison}
            \begin{tabular}{||l||c|c|c||}
                \toprule
                Tasks       & SPQ       & Odd speech & SRS \\
                \midrule
                \midrule
                Dream       & -0.24 & -0.24 & \textbf{-0.54} \\
                Favorite    & \textbf{-0.35} & -0.14 & \textbf{-0.27} \\
                Negative    & \textbf{0.22}  & \textbf{0.26} & \textbf{0.22} \\
                Mistake     & -0.07  & \textbf{0.18} & 0.15 \\
                \bottomrule
            \end{tabular}
        \end{table}
        Table~\ref{tab:task_comparison} shows Spearman's correlation between the ground truth and predicted values.  
        For all the target measures, the negative task showed a significant positive correlation. 
        In contrast, the favorite task corresponding to positive affects showed low prediction performance. 
        Although 
        some of them 
        showed significant negative correlations, we assume this is a pseudo correlation caused by poor prediction performance with nested cross-validation. 
        Similarly, Mota et al. reported better prediction performance with elicitation by negative images than positive ones, although they used different graph-based approaches \cite{Mota2017}. 
        A cognitive experiment by Minor et al. also showed that negative affect elicitation induced more positive FTD symptoms than the elicitation of neutral affects, negative affects, or cognitive load \cite{Minor2016}. 
        This result supports the usefulness of negative affect elicitation compared to other affective elicitation. 
        %
        %
        Compared to the negative task, the mistake task ineffectively elicited FTD symptoms, perhaps due to the perception uncertainty of the biggest mistakes. 
        Mistakes (even painful memories) sometimes lead to positive lessons; therefore the mistake task was ineffective in eliciting negative affects. 
        Therefore, the performance of the mistake task was lower than the negative task, which directly elicited negative affects. 
        
    \subsection{Duration effect}
    
    
        Although previous research used various speech durations to capture FTD symptoms, to the best of our knowledge, no study has tested its preferred duration. 
        Therefore, regarding RQ3, we compared three different speech durations: 30, 60, and 180 seconds. 
        Based on the result of the previous section, we used negative task data for this investigation. 
        
        \begin{table}[t]
            \centering
            \caption{
                \textbf{Total duration effect: Spearman's correlation between ground-truth and predictions:} Bold indicates significant correlations in the no-correlation test (p < 0.05).
            }
            \label{tab:total_duration_effect}
            \begin{tabular}{||l||c|c|c||}
                \toprule
                Tasks       & SPQ       & Odd speech & SRS \\
                \midrule
                \midrule
                30 sec.     & 0.08 & 0.17 & 0.30 \\
                60 sec.     & 0.20 & 0.21 & 0.07 \\
                180 sec.    & 0.06 & \textbf{0.37} & 0.19 \\
                \bottomrule
            \end{tabular}
        \end{table}
    
        Table~\ref{tab:total_duration_effect} shows the result of the model predictions. 
        In the \textit{odd speech} prediction, the correlation coefficients rose as the duration increased. 
        In contrast, we didn't find that tendency in the SPQ and SRS predictions. 
        Although we observed a correlation increase along with the duration, we failed to confirm whether the cause was the characteristics of the FTD or of the computational methods. 
        Hence, we divided the 180-second data into the following three phases (0-60, 60-120, and 120-180 seconds) and repeated the experiment. 
        \begin{table}[t]
            \centering
            \caption{
                \textbf{Speech phase effect: Spearman's correlation between ground-truth and predictions:} Bold indicates significant correlations in the no-correlation test (p < 0.05).
            }
            \label{tab:speech_phase_effect}
            \begin{tabular}{||l||c|c|c||}
                \toprule
                Tasks       & SPQ       & Odd speech & SRS \\
                \midrule
                \midrule
                30-60 sec.      & 0.09    & 0.03            & \textbf{0.36} \\
                60-120 sec.     & -0.02     & 0.14             & 0.02 \\
                120-180 sec.    & 0.20     & \textbf{0.40}    & 0.27 \\
                \bottomrule
            \end{tabular}
        \end{table}
        Table~\ref{tab:speech_phase_effect} shows the result of the prediction performance in the speech phases. 
        We confirmed a similar tendency in the \textit{odd speech}, which supported that the emergence of FTD characteristics induced an increase in the prediction performance. 
        %
        However, we confirmed correlation coefficients with SRS were higher at the beginning and the end, but in the middle lowered. 
        We assume this is related to attention deficits in ASD, reported difficulty ignoring distractions and maintaining attention for a long time \cite{Hogeveen2018}. 
        With the support of the previous experiment shown in Table~\ref{tab:total_duration_effect}; SRS prediction worked well in both short and long-duration settings, and the result suggests that longer speech duration is not required to elicit ASD-like symptoms. 
        
    \subsection{Ablation study}
    
        We ran an ablation study to compare the effectiveness of the features for the schizophrenic or autistic score prediction to answer RQ4. 
        Since we had to deal with several similar approaches, such as Tang's and Saga's methods, we grouped similar approaches and compared each group's effectiveness \cite{Tang2021,Saga2022_IEICE}. 
        As explained in Section~\ref{sec:ML_setting}, we divided the features into the following five groups: embedding, content word, function word, abstract, and temporal. 
        We excluded each feature group and evaluated the performance drop. 
    
        \begin{table}[t]
            \centering
            \caption{
                \textbf{Ablation study:} Bold indicates significant correlations in no-correlation test (p < 0.05).
            }
            \label{tab:ablation_study}
            \begin{tabular}{||l||c|c|c||}
                \toprule
                Eliminated group & SPQ & Odd speech & SRS \\
                \midrule
                \midrule
                All features    & \textbf{0.18} & \textbf{0.21} & \textbf{0.25} \\
                \midrule
                Embedding       & 0.12          & \textbf{0.19} & \textbf{0.24} \\
                Content word    & \textbf{0.17} & \textbf{0.23} & 0.18 \\
                Function word   & 0.09          & 0.04          & \textbf{0.09} \\
                Abstract        & \textbf{0.19} & \textbf{0.24} & \textbf{0.19} \\
                Temporal        & \textbf{0.19} & \textbf{0.17} & \textbf{0.28} \\
                \bottomrule
            \end{tabular}
        \end{table}
        
        Table~\ref{tab:ablation_study} shows the result of the ablation study. 
        Similar to the results in previous research, the function word group was the most effective feature for all score predictions.
        %
        For the ASD-related SRS prediction, in addition to the function words, content words worked as well. 
        As previous research reported people with ASD tend to use less complex language, so we assume that people with high ASD tendency characteristically use fewer modifiers (a part of content words) such as adjectives or adverbs compared to other people \cite{Eigsti2007}. 
        In contrast, the embedding-based features, which were used in a previous paper, showed a large drop only in the SPQ prediction. 
        This result supports Tang's finding of the relationship between embedding methods and Schizophrenia-related symptoms \cite{Tang2021}. 
        Furthermore, Saga reported that these embedding methods were significantly correlated with the human evaluations of word choice, fluency, and structure \cite{Saga2022_IEICE}. 
        Therefore, we assume these features captured the odd verbal discourse, including unusual phrase use or symptoms related to Schizophrenia characteristics \cite{Andreasen2016}. 
        Focusing on FTD-related symptoms with the \textit{odd speech} prediction, temporal features showed the effectiveness. 
        As mentioned in Andreasen's paper, FTD symptoms include \textit{poverty of speech}, a symptom of less speech amount. 
        Since temporal features measure those aspects, we assume this result supported that the features successfully captured poverty of speech \cite{Andreasen2016}. 
        However, further investigation is required to discover the mechanism of these phenomena from symptomatic perspectives. 
    
\section{Conclusion}

    This paper computationally investigated the differences between the linguistic characteristics of SPD and ASD. 
    We also compared the elicitation effects of speech tasks and speech duration settings.  
    Our correlation analysis between schizotypal and autistic measures suggested that a subscale of \textit{odd speech}, which is an FTD-related score, can bridge them (RQ1). 
    In terms of tasks and durations, we showed the effectiveness of negative memory reports with longer durations for eliciting FTD symptoms (RQ2, RQ3). 
    In addition, we investigated effective features to predict schizotypal and autistic scores, which revealed each linguistic characteristic (RQ4). 
    We assume that this investigation on linguistic characteristics of the symptoms is a part of digital phenotyping, which clarifies the schizophrenic and autistic languages with measurable computational methods. 
    Although most of the correlation seems weak, previous research showed similar findings described in section 3.2. 
    These findings support our result of the correlations. 
    We assume our methods contribute to future clinical diagnosis as supporting material with other multiple objective/subjective measures.
    %
    Concerning the future prospects of our approaches, they can be adopted for the computational verification of neurocognitive theories.
    Although researchers found the usefulness of NLP techniques in FTD characteristic investigation, its relative effectiveness was unknown. 
    Therefore, we compared them with the same dataset, which could help in the future verification of neurocognitive theories with the correspondence to FTD symptoms.
    In addition, the evaluation of our methods with schizophrenia or ASD participants is preferable for ensuring reliability and robustness. 

    
\begin{acks}

    Funding was provided by the Core Research for Evolutionary Science and Technology (Grant No. JPMJCR19A5), the Grant-in-Aid for Scientific Research (Grant No. 22K12151), the University fellowships toward the creation of science technology innovation (Grant No. JPMJFS2137), and the Grant-in-Aid for JSPS Research Fellow (Grant No. 23KJ1586).
    
\end{acks}

\clearpage
\balance

\bibliographystyle{ACM-Reference-Format}
\bibliography{ICMI2023}

\end{document}